\documentclass{IOS-Book-Article}

\usepackage[T1]{fontenc}
\usepackage[utf8]{inputenc}
\usepackage{url}
\usepackage{todonotes}
\usepackage{mathtools}
\usepackage{paralist}

\def\hb{\hbox to 10.7 cm{}}
\begin{document}
	\def\thepage{}
	
	\begin{frontmatter}

		\title{Leveraging Knowledge Graph Embedding Techniques for Industry 4.0 Use Cases}
		\runningtitle{KG Embedding for Industry 4.0}

		\author[A,C]{\fnms{Martina} \snm{Garofalo}},
		\author[A,C]{\fnms{Maria Angela} \snm{Pellegrino}},
		\author[B]{\fnms{Abdulrahman} \snm{Altabba}}, and
		\author[A,B,D]{\fnms{Michael} \snm{Cochez}
		\thanks{Corresponding Author: Michael Cochez; E-mail:
			\url{michael.cochez@fit.fraunhofer.de} other contacts:  \url{margar1994@gmail.com}, \url{mariaangelapellegrino94@gmail.com}, \url{abdulrahman.altabba@rwth-aachen.de}}}

		\runningauthor{Garofalo, Pellegrino, Altabba, and Cochez}
		\address[A]{Fraunhofer FIT, Sankt Augustin, Germany}
		\address[B]{Information Systems and Databases, RWTH Aachen University, Germany}
		\address[C]{Department of Computer Science,	University of Salerno, Italy}
		\address[D]{Faculty of Information Technology, University of Jyvaskyla, Finland}

\begin{abstract}
Industry is evolving towards \emph{Industry 4.0}, which holds the promise of increased flexibility in manufacturing, better quality and improved productivity.
A core actor of this growth is using sensors, which must capture data that can used in unforeseen ways to achieve a performance not achievable without them.
However, the complexity of this improved setting is much greater than what is currently used in practice. 
Hence, it is imperative that the management cannot only be performed by human labor force, but part of that will be done by automated algorithms instead.
A natural way to represent the data generated by this large amount of sensors, which are not acting measuring independent variables, and the interaction of the different devices is by using a graph data model.
Then, machine learning could be used to aid the Industry 4.0 system to, for example, perform predictive maintenance.
However, machine learning directly on graphs, needs feature engineering and has scalability issues.
In this paper we discuss methods to convert (embed) the graph in a vector space, such that it becomes feasible to use traditional machine learning methods for Industry 4.0 settings.

\end{abstract}
	\end{frontmatter}

\section{Introduction}
Industry 4.0 is a proposed set of ideas aimed at enhancing productivity and effectiveness by using a network of machines that cleverly collaborate.
To make the communication more intelligent new techniques must be applied, to succeed in helping human workers or, even, making the machines able to make their own decisions.
Artificial Intelligence, and in particular machine learning could be the core enabler of this automation.
Indeed, it has proven to enhance such fields as natural language processing and visual object recognition and surely could be the keystone either to improve the production process or to guarantee the quality of the realized work. 
Thanks to the Internet of Things (IoT) and Services in industrial processes, a large amount of data is generated daily and they can be represented in graph format.
Effective graph analytics provides users a deeper understanding of what is behind the data, and thus can benefit a lot of useful applications such as node classification, node recommendation, link prediction, etc.
However, most graph analytics methods suffer a high computation and space cost.
Moreover,  most algorithms work with a propositional \textit{feature vector} representation of the data, which means that each instance is represented as a vector of features, each of which are either binary, numerical or nominal.
In order to make industry data accessible to existing data mining tools, an initial \textit{propositionalization} of the corresponding graph is required.
Obviously the new set of propositional features could not encode all the knowledge available in the original data, but it can be effectively used to train a model via machine learning techniques and algorithms.
This could then be used to enhance the production process by, for example, monitoring how and which industrial components are used, detecting or predicting when a failure occurs, by checking the quality of the realized products, or by avoiding disastrous damage to the environment or to the factory taking into account the health of all the machineries.
Graph embedding is an effective yet efficient way to solve the graph propositionalization problem.
It converts the graph data into a low dimensional space in which the graph structural information and graph properties are maximally preserved.
The rest of the paper is structured as follows.
Section 2 presents an introduction to graph embeddings.
Section 3 explains some embeddings algorithms, with a focus on these usable for labeled directed graphs.
In section 4, we present the basic idea of industry 4.0 and how it could be enhanced by machine learning techniques enabled by embeddings.

\section{Graph Embeddings}
To get the best results for a given task, choosing the right embedding technique is obviously the starting point.
On the one hand, some of these techniques focus on representing specific parts of the graph (e.g., only nodes, also edges, or certain substructure) as vectors.
This typically happens by designing an objective function which optimizes for preservation of certain characteristics in the vector (like, for example, the edge reconstruction probability).
On the other hand, there are also some works concentrating on embedding complete graphs as vectors~\cite{DBLP:journals/corr/abs-1709-07604}. 
In addition to that, different types of graph input carry different information to be preserved in the embedded space and thus pose different challenges to the problem of graph embedding.
For example, when embedding a graph with node labels or edge attributes, the auxiliary informations provide interesting properties which cannot be seen otherwise, and thus may also be considered.
We divide the input graph into four categories, including homogeneous graph, heterogeneous graph, graph with auxiliary information, and graph constructed from non-relational data.
Moreover, we must keep in mind that the embedding output is task driven.
For example, the most common type of embedding output is a node embedding, with the aim of representing close nodes as similar vectors.
However, in some cases, the tasks may be related to higher granularity of a graph, e.g., node pairs, subgraph, whole graph.
Hence, the first challenge in terms of embedding output is to find a suitable embedding output type for the application of interest.
We categorize four types of graph embedding output, including node embedding, edge embedding, hybrid embedding, and whole-graph embedding.

\subsection{Graph types}
\textbf{Homogeneous graphs} (graphs with only one edge and node type) can be further categorized as weighted or unweighted  and directed or undirected graphs.
Since only structural information is available in homogeneous graphs, the challenge of homogeneous graph embedding lies in how to preserve these connectivity patterns observed in the input graph during embedding.
For \textbf{heterogeneous graph} the problem becomes assuring global consistency, preserving the different types related to the involved objects.
Heterogeneous graph embedding methods must be able to merge different types of objects, such as nodes and edges, into the same space .
The third category of input graph contains \textbf{auxiliary information} of nodes, edges or the whole-graph, in addition to the structural relations of nodes. 
Generally, there are different types of auxiliary information, such as labels and attributes.

In order to create an embedding, there are different techniques.
Commonly, embedding approaches will optimize an objective function, sometimes together with a classifier function.
Often also penalty on the similarity between nodes with different labels are introduced.
Another idea is to embed a knowledge graph, in which the entity (node) has a semantic category, perhaps in a hierarchical structure.
Sometimes, there is also a special treatment of attributes, which can have discrete or continuous values.
The goal is to incorporate that rich and unstructured information such that the embeddings are both representing the topological structure and discriminative in terms of the auxiliary information, preserving both of them.
The latter helps to define node similarity in addition to graph structural information. 
The first challenge faced by embedding graphs constructed from \emph{non-relational data} is how to compute the relations between the data and construct such a graph.
After the graph is constructed, the challenge becomes the same as in other input graphs, i.e., how to preserve the node proximity of the constructed graph in the embedded space.

\subsection{Embedding output}
Applications have different requirements regarding the information preserved in the embedding. 
Some embedding approaches aim at being general, meaning that they are suitable for a broad range of applications.
Other ones are created with a specific goal in mind.
In that case, finding the right approach is a challenging task.
Most commonly the output of an embedding approach is a so-called \textbf{node embedding}, which represents each node as a vector in low dimensional space.
Nodes that are \emph{`close'} in the graph are mapped to similar vector representations; the differences between various graph embedding methods lie in how they define the closeness between two nodes, and how they combine all the information contained in the input graph.
In contrast to a node embedding, an \textbf{edge embedding} represents each edge as a low dimensional vector.
In this case it is important to model the asymmetric property of the edges, if any. 
\textbf{Hybrid embedding }is the embedding of a combination of different types of graph components, e.g, nodes and edges (i.e., substructure) or nodes and community.
The last type of output is the \textbf{embedding of a whole graph},
Here, the key challenge is to satisfy the trade-off between the expressive power of learning complex properties and features, and the efficiency of the algorithm. 

\section{State of art}

According to Wang et al.~\cite{survey_approches_application}, a typical KG embedding technique generally consists of three steps: 
\begin{inparaenum}[1)]
	\item representing entities and relations,
	\item defining a scoring function, and
	\item learning entity and relation representations.
\end{inparaenum}
The first step specifies the form in which entities and relations are represented in a continuous vector space.
Entities are usually represented as vectors, i.e., deterministic points in the vector space. 
Relations are typically taken as operations in the vector space, which can be represented as vectors, matrices, tensors, multivariate Gaussian distributions or even mixtures of Gaussians.
Then, in the second step, a scoring function \textit{$f_{r}(h,t)$} is defined on each \textit{$fact(h,r,t)$} to measure its plausibility.
Facts observed in the KG tend to have higher scores than those that have not been observed.
Finally, to learn the entity and relation representations (i.e., embeddings), the third step solves an optimization problem that maximizes the total plausibility of observed facts.
In this overview, we focus on approaches which are designed to embed directed heterogeneous graphs which have types on both the edges and possibly on the nodes as well.

\subsection{TransE}
TransE \cite{transe} is an early and conceptually simple translational embedding model.
It represents both entities and relations as vectors in the same space; relationships are interpreted as translations, operating on the low-dimensional embeddings of the entities.
If in the modeled knowledge graph there is an edge \textit{l} connecting the entities \textit{h}(head) and \textit{t}(tail) then evaluating \textit{h + t} should produce a vector as close as possible to \textit{t}, additionally if there is no connection between \textit{h} and \textit{t} the result of \textit{h + l} should be far from \textit{t}. To achieve this characteristic we first assign random values to the entities and edges vectors, and then try to optimize them.
In other words we have to minimize the distance \textit{d(h + l, t)} if and only if \textit{(h, l, t)} holds.
To learn such embeddings, we minimize a margin-based ranking criterion over the training set:

\begin{equation}
\mathcal{L} = \sum_{(h,\ell,t)\in  S} \sum_{(h',\ell,t')\in S'_{(h,\ell,t)}}[\gamma + d(h+\ell,t)-d(h'+\ell,t')]_+ 
\end{equation}
where $[x]_+$ denotes the positive part of $x, \gamma>0$ is a margin hyperparameter, and 
\begin{equation}
S'_{(h,\ell,t)} = \left \{(h',\ell,t)|h' \in E \right \} \cup \left \{(h,\ell,t')|t' \in E \right \}
\end{equation}

Minimizing this loss function among all the existing entities and relationships by stochastic gradient descent, will shorten the distance between related entities more than other possible false relations with at least margin $\gamma$. An important condition that has to be added to the algorithm is that the      $\mathcal{L}_{2}$-norm of an entity vector has to be 1, while the relations vectors are unconstrained.
This condition is important as mentioned in [1] because it prevents the training process to trivially minimize $\mathcal{L}$ by artificially increasing entity embeddings norms.
Since the model is very simple, TransE is the most efficient algorithm.
Moreover, it has been proved to predict one-to-one link. 
However the corrupted triples used for link prediction testing are obtained by changing the head and the tail, one for time, of an actual triple.
In this way one corrupted triple could be a valid triple labeled wrongly.
The embedding vector for an entity is the same, no matter whether it appears first as a head and then as a tail in different triplets, in addition it uses the same space to model entities and predicate, as a result it can not deal with more complex properties, i.e., reflexive, one-to-many, many-to-one, many-to-many.

\subsection{TransH}
The aim of TransH is embedding a large scale knowledge graph into continuous vector spaces overcoming the problems pointed out against TransE and enabling an entity to have distributed representations when involved in different relations.
TransH \cite{transh} models a relation as a hyperplane together with a translation operation on it.
Each relation is characterized by two vectors: the norm vector of the hyperplane and the translation vector of the hyperplane.
For a golden triple, the projections of the head and the tail on the hyperplane are expected to be connected by the translation vector with low error.
The score function is 
\begin{equation}
f_{r}(h,t) = \left \| (h-w_{r}^{\top} h w_{r}) + d_{r}  - (t - w_{r}^{\top} t w_{r})\right \|_{2}^{2}
\end{equation}
where 
$h_{\perp} = (h-w_{r}^{\top} h w_{r}) + d_{r}$ (and consequently $t_{\perp} = (t-w_{r}^{\top} t w_{r}) + d_{r}$) by restricting $\left \| w_{r} \right \|_{2}=1$.
The function to minimize is:
\begin{equation}
\begin{multlined}
\mathcal{L} = \sum_{(h,r,t)\in \Delta } \sum_{(h',r',t')\in \Delta'_{(h,r,t)} }[f_{r}(h,t) +\gamma - f_{r'}(h',t')]_{+} \\
+ C\left \{ \sum_{e \in E } \left [ \left \| e \right \| _{2}^{2}-1 \right ]_{+} + \sum_{r\in R} \left [ \frac{(w_{r}^{T}d_{r})^{2}}{\left \| d_{r} \right \|_{2}^{2}} - \epsilon ^{2}\right ]_{+} \right \}
\end{multlined}
\end{equation}
where $[x]_+$ denotes the maximum between 0 and x; $\Delta$ is the set of golden triplets and $\Delta'$ the corrupted one; $\gamma$ is the margin separating positive and negative triplets; the function $f$ is the score function, lower for a golden triplet and higher otherwise and it takes into account the projections of node before comparing the head and the tail depending on the relationship between them; $C$ is the hyper-parameter weighting the importance of soft constraints;
where
\begin{equation}
\sum_{e \in E } \left [ \left \| e \right \| _{2}^{2}-1 \right ]_{+}
\end{equation}
models the constrains of the scaling : $\forall e \in E, \left \| e \right \| _{2} \leq 1$ and 
\begin{equation}
 \sum_{r\in R} \left [ \frac{(w_{r}^{T}d_{r})^{2}}{\left \| d_{r} \right \|_{2}^{2}} - \epsilon ^{2}\right ]_{+}
\end{equation}
models the constraints of the orthogonality, i.e. \\
$\forall r \in R, \frac{\left | w_{r}^{\top} d_{r}\right |}{\left \| d_{r} \right \|_{2}}\leq \epsilon  $. Moreover the constraint of the unit normal vector is satisfied projecting each normal vector to unit $\ell_{2}$-ball.

TransH is able to deal with reflexive, one-to-many, many-to-one, many-to-many properties and the corrupted triples for the testing phase are obtained by changing either the head or the tail of a golden triple with different probabilities depending on the mapping property of the relation, e.g. they give more chance to change the head if the relation is one-to-many.
The drawback of this model is the use of the same space to model entities and predicates.

\subsection{TransR}
TransR \cite{transr} shares a very similar idea with TransH, but it introduces relation-specific spaces, rather than hyperplanes.
For each triple $(h,r,t)$, entities in the entity space are first projected into $r$-relation space as $h_{r}$ and $t_{r}$ with operation $M_{r}$, and then $h_{r} + r \approx t_{r}$. The relation-specific projection can make the head/tail entities that actually hold the relation close with each other, and also get far away from those that do not hold the relation.
However this approach is not free from error since a specific relation, head-tail entity pairs usually exhibit diverse patterns.
For instance, the head-tail entities of the relation ``location-location\_contains" have many patterns such as country-city, country-university, continent-country and so on.
Following the idea of piecewise linear regression (Ritzema and others 1994), TransR was extended by clustering different head-tail entity pairs into groups and learning distinct relation vectors for each group, named as cluster-based TransR (CTransR). The training is done under the following objective function

\begin{equation}
L = \sum_{(h,r,t)\in  S} \sum_{(h',r,t')\in S'_{(h,\ell,t)}} \max (0,f_{r}(h,t) + \gamma - f_{r}(h',t'))
\end{equation}
where $\max(x,y)$ aims to get the maximum between $x$ and $y$, $\gamma$ is the margin, $S$ is the set of correct triples and $S'$ is the set of incorrect triples, $f_{r}(h,t)$, in the simplest form of TransR without clusters, is the score function 
\begin{equation}
f_{r}(h,t) = \left \| h_{\perp} + r - t_{\perp} \right \|_{2}^{2}
\end{equation}

The strength of TransR is dealing with fine grained models to handle complicated internal correlations under each relation type.
On the other hand, to achieve this, the computation complexity goes higher than both TransE and TransH. Moreover there are models more expressive than it.

\subsection{RDF2Vec}
RDF2Vec uses language modeling approaches for unsupervised feature extraction from sequences of words, and adapts them to RDF graphs.
The main idea is to adapt neural language models for RDF graph embeddings. 
Such approaches take advantage of the word order in text documents, explicitly modeling the assumption that closer words in a sequence are statistically more dependent.
In the case of RDF graphs, we consider entities and relations between entities instead of word sequences.
Thus, in order to apply such approaches on RDF graph data, we first have to transform the graph data into sequences of entities, which can be considered as sentences.
Using those sentences, we can train a neural language models to represent each entity in the RDF graph as a vector of numerical values in a latent feature space.
The algorithm \cite{rdf2vec} is divided into two steps: 1) convert the graph into a set of sequence of entities using two different approaches, i.e., Weisfeiler-Lehman Subtree RDF Graph Kernels and graph walks, 2) learn latent numerical representations of entities in RDF graphs through Word2vec, a particularly computationally-efficient two-layer neural net model. 
RDF2Vec model is evaluated on three different tasks: (i) standard machine-learning tasks (ii) entity and document modeling (iii) content-based recommender systems and for each of them reaches high level results, proving that this approach could be considered as task independent.
However, it is based on paths, walks, or kernels, and therefore rely on the identification of local patterns, this could be a limitation if we want a wider view to be captured by the embedding.
Moreover, techniques like graph kernels are not scalable for large dataset.

\subsubsection{Random graph walks}
For each vertex \textit{v} of the graph, they generate all graph walks of depth \textit{d} rooted in the vertex \textit{v} through breadth-first algorithm.
The final set of sequences for the given graph is the union of the sequences of all the vertices crossed by all the paths.~\cite{DBLP:conf/wims/CochezRPP17}

\subsubsection{Graph kernels}
The basic idea behind their computation is to evaluate the distance between two data instances by counting common substructures in the graph, i.e. walks, paths and trees.
The subtree RDF adaption is used.  The kernel computes the number of subtrees shared between two or more graphs by using the Weisfeiler-Lehman test of graph isomorphism.  To apply the algorithm, it must be adapted to the graph in input: it has to manage a directed edges and the labels from two iterations can potentially be different while still representing the same subtree.
To avoid it, at each iteration the neighboring labels of the previous iteration is compared to the actual one and, if they are identical, the label of the previous iteration is reused.
For each vertex, all the paths of depth \textit{d} within the subgraph of the vertex \textit{v} on the relabeled graph are extracted.
Then the original label of the vertex \textit{v} is set as the starting token of each path.
This process is repeat for \textit{h} times.
The final set of sequences is the union of the sequences of all the vertices in each iteration.

\subsubsection{Word2vec}
It is a one level neural networks for learning a low-dimensional and dense representation of words with two essential properties: (i) similar words are close in the vector space, and (ii) relations between pairs of words can be represented as vectors as well, allowing for arithmetic operations in the vector space.
Word2vec \cite{word2vec} estimates the likelihood of a sequence of entities appearing in the graph.
There are two different models: Continuous Bag-of-Words (CBOW) and Skip-Gram.
The CBOW model predicts target words from context words within a given window.
For each word of the vocabulary, the algorithm establishes which is the probability of the word being a target word.
The Skip-gram model does the inverse of the CBOW model and tries to predict the context words from the target words.
Hierarchical softmax and negative sampling are used as optimizations.
Once the training is finished, all the words (for instance the entities) are projected into a lower-dimensional feature space and semantically similar words (or entities) are positioned close to each other.

\subsection{Global RDF vector space embeddings}
KGloVe \cite{DBLP:conf/semweb/CochezRPP17}, similarly to RDF2Vec, takes advantage of the word embeddings, but using as an alternative, techniques able to identify global patterns, such as GloVe.
Glove training, however, is based on the creation of a global co-occurrence matrix from text.
Consequently, the first step is the building of a co-occurrence matrix from graph data.
To this end, they first weigh the edges of the graph and compute approximate personalized PageRank scores starting from each node.
The PageRank score for the other nodes (i.e., context nodes) is then used as the absolute frequency in the matrix.
This procedure is repeated on the graph with all edges reversed and the result is added to the co-occurrence matrix.
The obtained matrix is finally normalized.
This combined matrix is then subsequently used for training the vectors with the original GloVe approach.
In short, global RDF exploits global patterns for creating vector space embeddings, using 12 different weighting strategies to build the co-occurrence matrix. 
Besides, they propose many optimizations to speed up the computation of Personalized Page Rank.
Therefore this approach is able to incorporate larger portions of the graph, without substantially increasing the computational time of the RDF2Vec approach. 
In spite of this, the GloVe approach does not improve the embeddings created by RDF2Vec in the use cases tested in the paper.

\subsubsection{Co-occurrence matrix creation by Personalized PageRank}
The co-occurrence matrix for textual data is obtained by linearly scanning through the text and counting the occurrence of context words in the context of each word.
Instead the graph has not a linear structure.
To define it, the Personalized PageRank has been used.
It determines how important nodes in the context of a focus node.
The PageRank is used to find important nodes in a directed graph.
At its heart, it works by simulating random walkers over the graph and observing where these random walkers end up.
The simplified page rank problem is solved by finding the stationary solution to 
\begin{equation}
p^{(k+1)} = P^{T} p^{(k)}
\end{equation}
where P is a $n\times n$ matrix where n is the node number in the graph filled with zeros excepts for positions \textit{i,j} for which there exists an arc $ i \longrightarrow j $ and these positions contain $ \dfrac{1}{deg(i)} $ with \textit{deg(i)} is the out degree of the node \textit{i}.

To solve the problem of \textit{dangling nodes}, i.e. nodes with zero out degree, a node could continue from another node selected from a distribution \textit{v}, called the \textit{teleportation distribution}. Usually \textit{v} is chosen to be a uniform distribution.
To avoid ending in a cycle, a random jump is also performed with probability $\alpha$ to the focus node (this is a difference with the classical PageRank that consider the target node a node chosen with the same probability of all other nodes).
The Personalized PageRank assign a value to all nodes in the graph, ending up with a very large dense matrix with small values.
To make this step faster, the co-occurrence matrix is created by a fast personalized PageRank Approximation.

\subsubsection{Co-occurrence matrix creation by a fast personalized PageRank Approximation}
The effort of the algorithm is only used for the nodes which will receive a significant rank. \textit{b} is the focus node and the algorithm starts injecting a unit amount of paint to \textit{b}. The paint represents the random walk in the original algorithm.
From this paint, an $\alpha$-portion is retained and added to the value for \textit{b} in $p^{(b)}$. The remaining (1-$\alpha$)-portion is distributed uniformly over the out-links.
This retain-and-distributed process is repeated recursively for all nodes.
When a node has a zero out degree, the outgoing paint is discarder.
The best performance is obtaining by choosing the oder in which nodes are evaluated, reusing the value of the vector \textit{p}, and changing the uniform distribution with a biased one.

\subsubsection{The GloVe model}
GloVe is designed for creating word embeddings from natural language texts.
It creates the co-occurrence matrix evaluating the size of the context window, distinguishing left from right context and a weighting function to weight the contribution of each word co-occurrence.
Then, it tries to minimize the following cost function:
\begin{equation}
J = \sum_{i,j = 1}^{V} f(X_{ij}) (w_{i}^{T}\cdot\widetilde w_{j}+b_{i}+\widetilde b_{j}-logX_{ij})^{2}
\end{equation}
where $f(X_{ij})$ is a weighting function on co-occurrence counts of word \textit{j} in the context of word \textit{i}, value represented by $X_{ij}$, 
$w_{i}$ are word vectors, $\widetilde w_{j}$ are context vectors, $ b_{i} $ and $\widetilde b_{j} $ are biases.
The idea is that when two words co-occur often, their vectors' dot product will be relatively high, meaning that the vectors are more similar to make the error factor smaller.

\section{Industry 4.0 contextualization}

Kagermann and Wahlster in \cite{industryDef} characterized Industry 4.0 as follows:
\begin{quote}
	``Industry 4.0 will involve the technical integration of CPS [cyber-physical systems] into manufacturing and logistics and the use of the Internet of Things and Services in industrial processes.''
\end{quote}

It is expected that the Industry 4.0 era, where intelligent analytics and cyber-physical systems team together, will realize a new thinking of production management and factory transformation; all objects of the factory world would be equipped with integrated processing and communication capabilities.
This change does not only affect machine to machine (M2M) communication, but will also have far-reaching consequences for the interplay of humans and technology.
Industry 4.0 is not geared towards worker-less production facilities.
Instead, people should be integrated into the cyber-physical structure in such a way that their individual skills and talents can be fully realized.
The interplay between human and CPS occurs either by direct manipulation, or with the help of a mediating user interface.
The primary function of the worker will thus be to dictate a production strategy and supervise the implementation, which is carried out by self-organizing production processes.
In production environment, data is generated and collected from different machines and environment sensors, processes, products, quality indicators, logistics , external partners, and infrastructure; all contribute into explosion in data size.
This amalgamation of data is often referred to as \emph{Big Data}.

The growth of data fires the necessity to introduce new techniques to enable exploitation by extracting new knowledge and information.
One group of techniques is machine learning, which has become the main driver of innovation in certain industrial sectors.
In this section we will discuss some possible applications for machine learning in industry 4.0, which can be tapped into using the embedding techniques discussed in the previous sections.
We will mainly focus on two aspects which machine learning can bring into the production processes, namely a) predictive maintenance to anticipate machine breakdowns and b) improvement of product quality.

\subsection{Predictive maintenance}
A well-established and relatively simple method of recognizing failures early on, is condition monitoring.
The complexity of forecasting failure is often due to the enormous amount of possible influencing factors, furthermore data sources can be manifold and depend on the scenario.
AI-based algorithms are capable of recognizing errors and differentiating the noise from the important information to predict breakdowns and guide future decisions even in complex scenario.
The workflow is usually the following: the sensors generate data which is then compared to the information from the machine and the specific workpiece being processed.
By analyzing the data, it is possible to identify patterns of behavior that predict when a problem is occurring with high accuracy.
This enables maintenance schedules to be planned accordingly.
Briefly the approach is the following: some machine-learning techniques examine the relationship between a data record and the labeled output (e.g., failures) and then create a data-driven model to predict those outcomes.
These techniques help to recognize patterns from historical events and either predict future failures, or prevent them based on learnings from specific breakdown root causes. 
The input graph could change based on the failure you want to identify: 
\begin{description}
	\item[node failure] the graph could model the status of the node or the amount of energy produced by it.
	In the first case all the nodes are equals and you have to model their values by a binary function, otherwise you need to use a numerical function to model their activity.
	The embedding could ignore edges and it can focus on the nodes embedding.
	\item[edge failure] the failure could be modelled by error in communication among nodes.
	In this case the input graph must model the interactions either by a binary function (there is exchange of data, or not) or by a numerical function (to mean the amount of information exchanged). To this end, embedding techniques able to manage both nodes and edges must be used, even without a global view.
	\item[subgraph failure] sometimes for each node there is a range of accepted values.
	Even if the actual nodes values are in these windows, looking at the whole picture you can see a failure arise caused by a combination of factors.
	In this case the used embedding technique must be able to model subgraphs or inner structures.
	\item[graph failure] is a special type of subgraph failure where the whole system must be monitored.
	In this case the embedding technique must be able to model the whole graph with a global view.
\end{description}
Starting from the input graph that model the system to monitor, the chosen embedding technique converts it in vectors which can be used by the machine-learning algorithms to recognize either failures or correct behaviour.
It is important to underline the fact that most of the embedding techniques showed till now do not consider literals in their models, so if the graph encodes significant information in such a way, it is a key aspect to take into account when choosing the embedding strategy.
In order to estimate the impact on system performance, two key factors need to be considered according to \cite{self-maintenance}: (i) the mitigation of production uncertainties to reduce unscheduled downtime and increase operational efficiency, and (ii) the efficient utilization of the finite resources on the critical sections of the system by detecting its bottleneck components.
Comparing an AI-based approach to traditional condition monitoring or more classical maintenance strategies like usage-based exchange, a considerable improvement can be expected due to better failure prediction.
Depending on the starting point and the level of redundancy, availability can sometimes increase by more than 20\%. Inspection costs may be reduced by up to 25\% and an overall reduction of up to 10\% of annual maintenance costs is possible~\cite{mckinsey}.
\subsubsection{Use cases} 
We can apply this general idea to more specific use cases:
\begin{description}
	\item[Predictive maintenance for heat exchangers]
	Deposits in conduits can cause heat exchangers to clog.
	A complete blockage of such system can cause serious problems, resulting in manufacturing errors and hours of downtime.
	Ideally, it would be possible to measure that this is happening and take countermeasures, but this is complicated by the fact that it is impossible to measure the flow rate of a heat exchanger directly.	
	One solution to this issue is to measure the temperature differential upstream and downstream of the heat exchanger.
	Gathering the measured values for a certain time span, it is possible to create a profile for normal behavior and define threshold values.	
	These models can then be used as input for an alert system to notify as soon as the first signs of clogging appear.
	To help the monitoring of the system, various visualization techniques can be applied.
	As this approach helps to efficiently recognize anomalies that could cause potential blockages, machine downtime is reduced and also other devices connected to the heat exchanger will be less strained.
	
	\item[Predictive maintenance for climate and energy]
	Climate change and energy transition are two of the biggest challenges facing us today.
	Industry, especially the players in the energy market, must deal with the consequences of this shift. 
	Thanks to machine learning techniques, they can manage the increasingly complex market, by attempting to match demand and supply.  
	In fact, based on historical energy consumption patterns, the expected demand can be derived, and an intelligent control system can ensure a price-optimized strategy for power generation in real-time.

	\item[Predictive maintenance for the health of robots]
	It is difficult to plan robot maintenance if the health of a robot is monitored only locally or not at all.
	Many parameters can be monitored, including CPU and housing temperature, as well as positioning and overload errors.
	Doing so makes intervention before the machine is damaged possible. 
	Besides, understanding the root cause of the problem makes predictive maintenance possible, increasing uptime.

\end{description}
	
\subsection{Quality control}
Particularly in Germany, the quality of artifacts realized in the context of industrial production is of outstanding importance.
However, in the traditional workflow, the quality of the products is only checked at the end of the realization process.
Machine Learning turns this relationship upside down: the use of sensor technology and the continuous evaluation of data at component level makes it possible to check and ensure the state of parts during operation.
Particularly when sources of error can be determined beforehand and their variables influenced, individual measurement data collected during production and tests can be integrated into the production process.
Therefore, a test automation based on machine learning can significantly improve the throughput of production. 

To apply these changes, we notice that, from an abstract level, there is similarity between these use cases and the previously analyzed sub-graph failure.
The graph should be modeled such that interactions among the different components that play a role in the process are captured and the embedding technique must represent how the flow proceeds, taking into account extra information measured by sensor networks and related to values to monitor.
There could be patterns that are always wrong or some paths that are correct in some cases and wrong otherwise.
For this reason, based on the product we are creating, different patterns could be recognized and classified as correct or not.
The same reasoning could be applied to monitored values: in some circumstances, observations are justified while in other cases they are suspicious or incorrect.
When the input graph is modeled, one has to pay carefully combine elements produced by different parts of the sensor network.
One issue that can emerge is that devices created by different vendors, model data with incompatible formats.
A solution to this could be to model, besides sensors and data, information on how to convert them or how to create relationships among data directly into the embedding approach.
To estimating the impact of such changes on the quality control, the main success factors are related to avoidance of costs associated with shipping bad products, the accuracy of error detection, a better productivity, lower deployment time, and improvement of efficiency and speed since the need for human input is lowered significantly.
As reported in a recent McKinsey report \cite{mckinsey}, the advancements in AI-based quality assurance promise productivity increases of up to 50\% and defect detection using deep-learning-based systems could improve up to 90\% in comparison to human inspection.
To sum up, generally, insights from AI-based quality testing can be used for root cause analysis to improve the overall production process.

\subsubsection{Use cases} 
Depending on the industrial field, there could be different kinds of sensors to monitor the workflow:
\begin{itemize}
	\item in \textbf{automotive} factories there is the need of monitoring high-level procedures, such as welding, metal finishes or coatings, all using sensors of precision, positioning, high temperature or pressure;
	\item in \textbf{engineering} factories the focus is on temperature, electromagnetic interference, cutting and bonding;
	\item in \textbf{textile} field could be useful monitoring the precision of the realized products and taking into account the timing of the operations; and
	\item in fields where \textbf{filling} machines are used, there is the need of sensors able to detect containers, covers, filling levels and labels, checking everything accurately.
\end{itemize}

\subsection{Context-aware robots}
Industrial robots are often special-purpose and they are not able to react to changes in their environment.
For these reason, it is better to fence them and let them work in robot-only-area.
Nowadays, AI enhancement are enabling a new generation of automation robots, developing them in such way that their responses can adapt flexibly to any changes of the context, and so they are not special-purpose anymore.
The advantages are less configuration time and adaptability to specific environments, enabling humans and robots to work together, in the same space.
This is an important shift since traditional machines do not come out of theirs predefined steps, thus it is not possible to exploit other knowledge data sources to react to unusual situations.
Sometimes the collaboration between machine and human could be the way through which the robots learn new actions, for instance a factory could use AI to ``program'' a robot by simply showing the desired movements to it.
So the robot ability is the result of a combination between predefined rules and freshly leaned movements.
This knowledge could then also be exploited for training new employees, overturning the learning process. 

The first step towards context-aware robots, is gathering data, such as the device status,
production demand information and environmental parameters.
Then, the data can be analyzed, and context-aware services may be realized.
The second step is designing optimized decision-making algorithms under certain constraint conditions, such as energy efficiency or cost savings.
The focal point of this approach is that all the robotics and other smart devices mutually collaborate and form a computing resource network, which provides a pool of shared computation and storage resources, elastically allocated according to real-time requirements.
Through the cloud, information can be exploited to form a new intelligent system: 
\begin{inparaenum}[1)]
	\item all the data from the devices and environments are analyzed to provide some novel services,
	\item the skills or behaviors of cloud robots may form a knowledge library containing information which is shared among the cloud robots for learning purposes.
\end{inparaenum}

McKinsey Global Institute (MGI) estimates an automation potential across the German manufacturing sector of 55\% of currently performed activities, freeing up capacity to focus on value-added work.
Employing context-aware robots in logistics processes related to the delivery of product parts, e.g., is expected to produce efficiency gains of 5 to 10\% in picking and a 15 to 20\% reduction in travel time.
Collaborative robots are particularly relevant with respect to tasks that are not fully automated.
In such settings they hold the potential of increasing productivity by up to 20\%. 
The implementation of flexible and adaptable robots will change the way manufacturing processes are set up.
Manufacturing lines will look more like flexible work aisles where humans and robots work together side by side and an increasing number of tasks are performed by one system \cite{mckinsey}.	

\subsubsection{Use cases}
Context-aware robots could be useful in decision-making mechanisms.
They can be developed depending on the goal to achieve or the factor to minimize.
For example, they can be exploited in 
\begin{description}
	\item[Energy efficiency scheme] for material handling: the goal is to find the shortest path in a limited robotic resources environment.
	In this case information interaction and sharing between manipulation platforms and robots may be realized by designing a scheduler.
	This scheduler seeks the minimum element of stack vectors that reflects the requirements for material handling, and analyzes the location of all the robots.
Once the robot and the platform are determined, the scheduler redirects the target robot and makes it move to the right platform to unload the workpiece~\cite{context-awareRobots}. 
\end{description}

\subsection{Marketing strategy}
To stay competitive, companies are always looking at cost reductions which do not lower the quality of their products (or rather, savings which do not cause customer dissatisfaction).
As reported in~\cite{ smartDesign}, physical prototyping is being replaced by virtual processes since multiple configurations can be compared and evaluated, fewer materials are needed, and it is more cost and time efficient.
Virtual prototyping makes use of specific software to simulate and validate the performance of a design concept before an actual physical prototype is build.
Then, engineers have the luxury of exploring many design alternatives before investing the time and money to construct physical prototypes.
In summary, the benefits of this method is a reduced time-to-market and cost, as well as products with higher level of performance and reliability.

When designing a new product, the aim is to find the best design within known range, making an optimization of a known process.
On the contrary we could find new and better design through learning, creating something unusual.
Moreover, it is possible to consider an adaptation of the work, based on what users want, as is explained in the work by Saldivar et al.~\cite{customDesign}.
There are different applications in which the user desiderata are at the center of attention, starting from the automotive design \cite{customDesign} to the tourist guide personalization~\cite{tourPersonalization}. 
In the latter reference, Knowledge Graphs are used to personalize the guide in museums based on the customer interest and location.
Besides the guide will adapt when the user changes his mind and deviates from the original computed path, the Knowledge Graph could be used to inspire new products and modification to apply to the current products, based on what the public like.
These ideas could be used both to change the style of the production and surprise the customers, or to satisfy special orders for a limited audience.

From a technical point of view, during the creation phase, suggestion from the knowledge graph can be exploited, exactly as in a recommendation system.
The input graph models all the features which are interesting for customers and represent how they can be put together.
For instance, it could model that people who like blue are interested in astronomy or stars.
These features can be embedded in a feature vector representation and used for a classification algorithm.
In this way the inventor could use these related properties to create some coherent and consistent design based on customers interests.
Otherwise, the information could be used as a dissimilarity measure to embed the graph information.
In this way the designer could put together very dissimilar features to create something original, or propose very different designs aiming at filtering out options as efficiently as possible.
The \textit{modus operandi} depends on the target of the product or the style that the factory is used to follow.

\subsubsection{Use cases}
Virtual prototyping has become an important part of the industry workflow in many fields:
\begin{itemize}
	\item in \textbf{automotive} factories it is usual to prototype new cars and test them in advanced simulation environments to try out performance or compare the aesthetics with the actual trends to achieve the best results;
	\item in general every factory could exploit this method to guide its own industry strategy.
For instance by merging information from different knowledge base such as shipping trades per product type and factory distribution all over the globe, it could be recommend a strategic place where a new branch can be built.
	Moreover, by using virtual prototyping in addition to well known sales information, the push towards a new model of marketing seems obvious. 
\end{itemize}  

\section{Conclusions}
On the one hand in the Industry 4.0 era, working processes are enhanced by sensor networks that cleverly collaborate and which interactions or features could be modeled by graph.
On the other hand AI is mainly targeted towards recognizing patterns and anomalies in feature vectors containing binary, numerical or nominal values. 
In order to make industry data accessible to existing data mining tools, an initial propositionalization of the corresponding graph is required.
This bridge could be achieved by using embedding techniques which convert the input graph in numerical vectors.
The data produced by the sensor networks could, after an optional cleaning process, be embedded in numerical feature vectors to train machine learning models.
These models could then be used in the industry to enhance productivity, improve the quality of the production, foresee machine breaking, and seem a step towards reaching the coveted phase called Industry 4.0. 

\section*{Acknowledgment}
We would like to thank the participant of the NATO Advanced Research Workshop -- G5172 ``Cyber Defence in Industry 4.0 Systems and Related Logistics and IT Infrastructures'' for the feedback and input received during and after the event.
Part of this work was created as part of the Linked Data seminar at the I5 chair of the RWTH university.

\bibliographystyle{asmems4}
\bibliography{asme2e}

\begin{thebibliography}{10}

\bibitem{DBLP:journals/corr/abs-1709-07604}
Cai, H., Zheng, V.~W., and Chang, K.~C., 2017.
\newblock ``A comprehensive survey of graph embedding: Problems, techniques and
  applications''.
\newblock {\em CoRR, {\bf abs/1709.07604}}.

\bibitem{survey_approches_application}
Wang, Q., Mao, Z., Wang, B., and Guo, L., 2017.
\newblock ``Knowledge graph embedding: A survey of approaches and
  applications''.
\newblock {\em IEEE Transactions on Knowledge and Data Engineering, {\bf
  29}}(12), Dec, pp.~2724--2743.

\bibitem{transe}
Bordes, A., Usunier, N., Garcia-Duran, A., Weston, J., and Yakhnenko, O., 2013.
\newblock ``Translating embeddings for modeling multi-relational data''.
\newblock In {\em Advances in Neural Information Processing Systems 26},
  C.~J.~C. Burges, L.~Bottou, M.~Welling, Z.~Ghahramani, and K.~Q. Weinberger,
  eds. Curran Associates, Inc., pp.~2787--2795.

\bibitem{transh}
Wang, Z., Zhang, J., Feng, J., and Chen, Z., 2014.
\newblock ``Knowledge graph embedding by translating on hyperplanes''.
\newblock {\em AAAI Conference on Artificial Intelligence}.

\bibitem{transr}
Lin, Y., Liu, Z., Sun, M., Liu, Y., and Zhu, X., 2015.
\newblock ``Learning entity and relation embeddings for knowledge graph
  completion''.
\newblock In Proceedings of AAAI'15.

\bibitem{rdf2vec}
Ristoski, P., Rosati, J., Di~Noia, T., De~Leone, R., and Paulheim, H., 2016.
\newblock ``{RDF2Vec}: {RDF} graph embeddings and their applications''.
\newblock {\em Semantic Web journal}.

\bibitem{DBLP:conf/wims/CochezRPP17}
Cochez, M., Ristoski, P., Ponzetto, S.~P., and Paulheim, H., 2017.
\newblock ``Biased graph walks for {RDF} graph embeddings''.
\newblock In Proceedings of the 7th International Conference on Web
  Intelligence, Mining and Semantics, WIMS 2017, Amantea, Italy, June 19-22,
  2017, R.~Akerkar, A.~Cuzzocrea, J.~Cao, and M.-S. Hacid, eds., ACM,
  pp.~21:1--21:12.

\bibitem{word2vec}
Mikolov, T., Chen, K., Corrado, G., and Dean, J., 2013.
\newblock ``Efficient estimation of word representations in vector space''.
\newblock {\em CoRR, {\bf abs/1301.3781}}.

\bibitem{DBLP:conf/semweb/CochezRPP17}
Cochez, M., Ristoski, P., Ponzetto, S.~P., and Paulheim, H., 2017.
\newblock ``Global {RDF} vector space embeddings''.
\newblock In The Semantic Web - {ISWC} 2017 - 16th International Semantic Web
  Conference, Vienna, Austria, October 21-25, 2017, Proceedings, Part {I},
  C.~d'Amato, M.~Fern{\'{a}}ndez, V.~A.~M. Tamma, F.~L{\'{e}}cu{\'{e}},
  P.~Cudr{\'{e}}{-}Mauroux, J.~F. Sequeda, C.~Lange, and J.~Heflin, eds.,
  Vol.~10587 of {\em Lecture Notes in Computer Science}, Springer,
  pp.~190--207.

\bibitem{industryDef}
Kagermann, H., Helbig, J., Hellinger, A., and Wahlster, H., 2013.
\newblock ``Recommendations for implementing the strategic initiative
  {Industrie} 4.0: Securing the future of {German} manufacturing industry''.
\newblock {\em Final report of the {Industrie} 4.0 Working Group}.

\bibitem{self-maintenance}
Lee, J., Kao, H.-A., and Yang, S., 2014.
\newblock ``Service innovation and smart analytics for {Industry} 4.0 and {Big
  Data} environment''.
\newblock {\em Procedia CIRP, {\bf 16}}, pp.~3 -- 8.
\newblock Product Services Systems and Value Creation. Proceedings of the 6th
  CIRP Conference on Industrial Product-Service Systems.

\bibitem{mckinsey}
McKinsey, 2017.
\newblock ``Smartening up with artificial intelligence ({AI}) - what's in it
  for {Germany} and its industrial sector?''.

\bibitem{context-awareRobots}
Wan, J., Tang, S., Hua, Q., Li, D., Liu, C., and Lloret, J., 2017.
\newblock ``Context-aware cloud robotics for material handling in cognitive
  industrial internet of things''.
\newblock {\em IEEE Internet of Things Journal}.

\bibitem{smartDesign}
Ang, J.~H., Goh, C., and Li, Y., 2016.
\newblock ``Smart design for ships in a smart product through-life and
  {Industry} 4.0 environment''.
\newblock In 2016 IEEE Congress on Evolutionary Computation (CEC),
  pp.~5301--5308.

\bibitem{customDesign}
Saldivar, A. A.~F., Goh, C., n.~Chen, W., and Li, Y., 2016.
\newblock ``Self-organizing tool for smart design with predictive customer
  needs and wants to realize {Industry} 4.0''.
\newblock In 2016 IEEE Congress on Evolutionary Computation (CEC),
  pp.~5317--5324.

\bibitem{tourPersonalization}
van Hage, W.~R., Stash, N., Wang, Y., and Aroyo, L., 2010.
\newblock ``Finding your way through the {Rijksmuseum} with an adaptive mobile
  museum guide''.
\newblock In The Semantic Web: Research and Applications, L.~Aroyo,
  G.~Antoniou, E.~Hyv{\"o}nen, A.~ten Teije, H.~Stuckenschmidt, L.~Cabral, and
  T.~Tudorache, eds., Springer Berlin Heidelberg, pp.~46--59.

\end{thebibliography}

\end{document}